%% file: main_cvpr.tex
\documentclass[10pt,twocolumn,letterpaper]{article}

\usepackage{cvpr}
\usepackage{times}
\usepackage{graphicx}
\usepackage{amsmath}
\usepackage{amssymb}
\usepackage[normalem]{ulem}

\usepackage{authblk}

\usepackage[pagebackref=true,breaklinks=true,letterpaper=true,colorlinks,bookmarks=false]{hyperref}

\cvprfinalcopy 
\def\cvprPaperID{3141} 

\ifcvprfinal\fi

\begin{document}
	

\title{
	\vspace{-8mm}
	DeepCap: Monocular Human Performance Capture Using Weak Supervision
	}

\author[1,2]{\vspace{-4mm}Marc Habermann}
\author[1,2]{Weipeng Xu}
\author[3]{Michael Zollhoefer}
\author[1,2]{Gerard Pons-Moll}
\author[1,2]{Christian Theobalt}
\affil[ ]{
\vspace{-5mm}\textsuperscript{1}Max Planck Institute for Informatics,
\textsuperscript{2}Saarland Informatics Campus,
\textsuperscript{3}Stanford University
}

\maketitle


\begin{abstract}
\input{sec/abstract}
\end{abstract}


\input{sec/intro.tex}
\input{sec/relatedwork.tex}
\input{sec/method.tex}
\input{sec/evaluation.tex}
\input{sec/conclusion.tex}

\par \noindent \textbf{Acknowledgements.}
This work was funded by the ERC Consolidator Grant 4DRepLy (770784) and the Deutsche Forschungsgemeinschaft (Project Nr. 409792180, Emmy Noether Programme, project: Real Virtual Humans).
%

{\small
\bibliographystyle{ieee}
\bibliography{bibliography}
}


\end{document}

%% file: sec/abstract.tex
\label{sec:abstract}
Human performance capture is a highly important computer vision problem with many applications in movie production and virtual/augmented reality.
Many previous performance capture approaches either required expensive multi-view setups or did not recover dense space-time coherent geometry with frame-to-frame correspondences.
We propose a novel deep learning approach for monocular dense human performance capture.
Our method is trained in a weakly supervised manner based on multi-view supervision completely removing the need for training data with 3D ground truth annotations.
The network architecture is based on two separate networks that disentangle the task into a pose estimation and a non-rigid surface deformation step.
Extensive qualitative and quantitative evaluations show that our approach outperforms the state of the art in terms of quality and robustness.

%% file: sec/intro.tex
%
\vspace{-4mm}
\section{Introduction}
\label{sec:intro}
%
%
\par
Human performance capture, i.e, the space-time coherent 4D capture of full pose and non-rigid surface deformation of people in general clothing, revolutionized the film and gaming industry in recent years.
Apart from visual effects, it has many use cases in generating personalized dynamic virtual avatars for telepresence, virtual try-on, mixed reality, and many other areas.
In particular for the latter applications, being able to performance capture humans from \emph{monocular video} would be a game changer.
The majority of established monocular methods only captures articulated motion (including hands or sparse facial expression at most). 
However, the monocular tracking of dense full-body deformations of skin and clothing, in addition to articulated pose, which play an important role in producing realistic virtual characters, is still at its infancy.
%
%
\par
In literature, multi-view marker-less methods \cite{bray06,brox06,brox10,cagniart10,de08,gall09,liu11,mustafa15,vlasic08,vlasic09, wu13,pons17,pons15} have shown compelling results. 
However, these approaches rely on well-controlled multi-camera studios (typically with green screen), which prohibits them from being used for location shootings of films and telepresence in living spaces. 
%
%
\begin{figure}[t]
	\begin{center}
		\includegraphics[width=\linewidth]{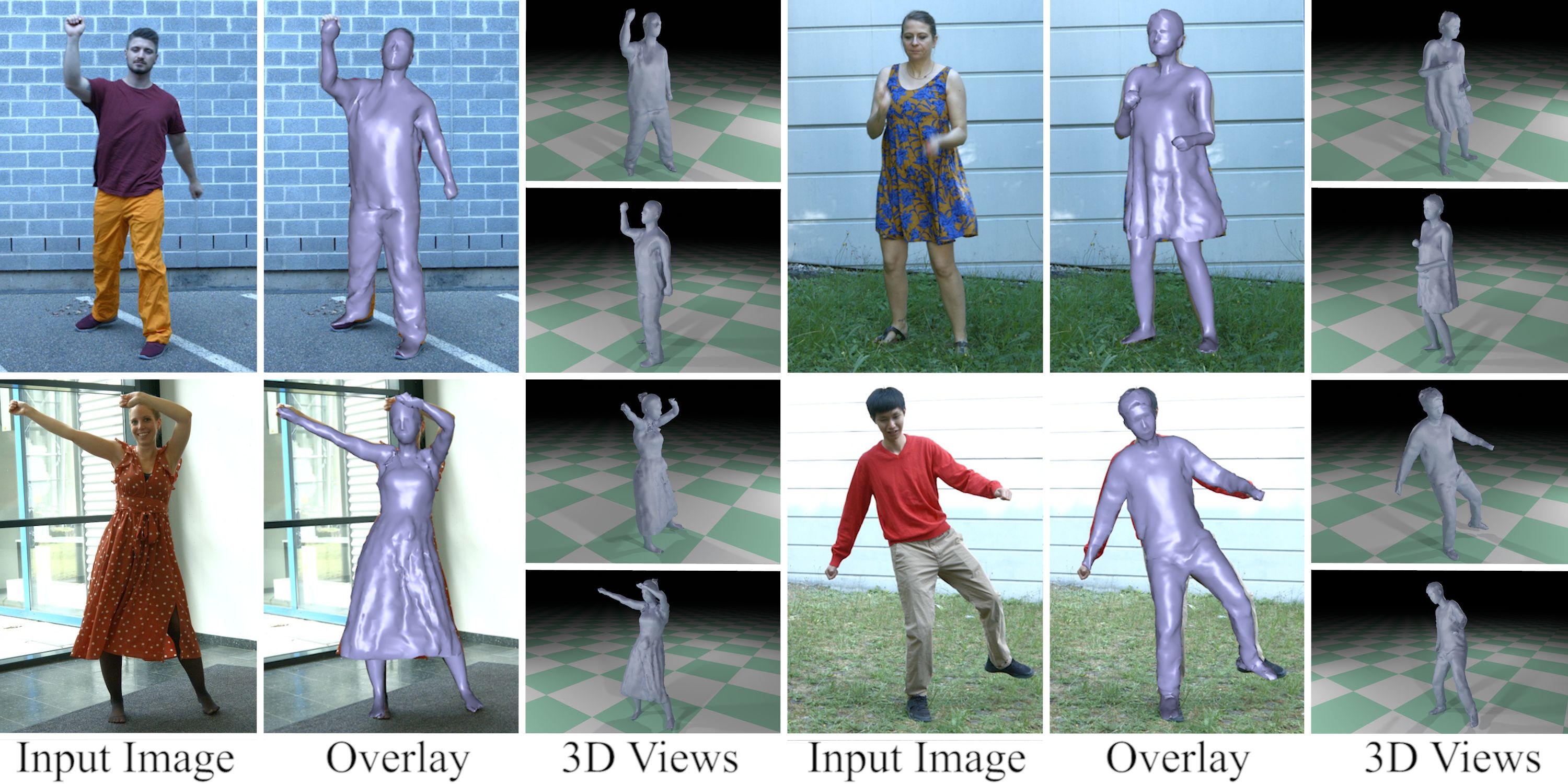}
	\end{center}
	\vspace{-2mm}
	\caption
	{
		We present the first learning-based approach for dense monocular human performance capture using weak multi-view supervision that not only predicts the pose but also the space-time coherent non-rigid deformations of the model surface.
	}
	\label{fig:teaser}
	\vspace{-5mm}
\end{figure}
\par
Recent monocular human modeling approaches have shown compelling reconstructions of humans, including clothing, hair and facial details~\cite{saito19,zheng19,alldieck19,alldieck18a,bhatnagar2019mgn, patel20, ma20}.
Some directly regress voxels~\cite{gabeur2019moulding,zheng19} or the continuous occupancy of the surface~\cite{saito19}. 
Since predictions are pixel aligned, reconstructions have nice detail, but limbs are often missing, especially for difficult poses. 
Moreover, the recovered motion is not factorized into articulation and non-rigid deformation, which prevents the computer-graphics style control over the reconstructions that is required in many of the aforementioned applications. 
Importantly, surface vertices are not tracked over time, so no space-time coherent model is captured.
Another line of work predicts deformations or displacements to an articulated template, which prevents missing limbs and allows more control~\cite{alldieck19,bhatnagar2019mgn,alldieck2019tex2shape,Pumarola_2019_ICCV}.
However, these works do not capture motion and the surface deformations. 
%
%
\par
The state-of-the-art monocular human performance capture methods~\cite{xu18,habermann19} densely track the deformation of the surface.
They leverage deep learning-based sparse keypoint detections and perform an expensive template fitting afterwards.
In consequence, they can only non-rigidly fit to the input view and suffer from instability.
By contrast, we present the first learning-based method that jointly infers the articulated and non-rigid 3D deformation parameters in a single feed-forward pass at much higher performance, accuracy and robustness.
The core of our method is a CNN model which integrates a fully differentiable \emph{mesh} template parameterized with \emph{pose} and an \emph{embedded deformation graph}. 
From a single image, our network predicts the skeletal pose, and the rotation and translation parameters for each node in the deformation graph. 
In stark contrast to implicit representations~\cite{saito19, zheng19,chibane20ifnet}, our mesh-based method \emph{tracks the surface vertices over time}, which is crucial for adding semantics, and for texturing and rendering in graphics. 
Further, by virtue of our parameterization, our model always produces a human surface \emph{without missing limbs}, even during occlusions and out-of-plane motions. 
%
%
\par
While previous methods~\cite{saito19,zheng19,alldieck19,bhatnagar2019mgn} rely on 3D ground truth for training, our method is weakly supervised from multi-view images.
To this end, we propose a fully differentiable architecture which is trained in an analysis-by-synthesis fashion, without explicitly using any 3D ground truth annotation.
Specifically, during training, our method only requires a personalized template mesh of the actor and a multi-view video sequence of the actor performing various motions.
Then, our network learns to predict 3D pose and dense non-rigidly deformed surface shape by comparing its single image feed-forward predictions in a differentiable manner against the multi-view 2D observations.
At test time, our method only requires a single-view image as input and produces a deformed template matching the actor's non-rigid motion in the image.
In summary, the main technical contributions of our work are:
\begin{itemize}
	\item{A learning-based 3D human performance capture approach that jointly tracks the skeletal pose and the non-rigid surface deformations from monocular images.}
	\item{A new differentiable representation of deforming human surfaces which enables training from multi-view video footage directly.}
\end{itemize}
Our new model achieves high quality dense human performance capture results on our new challenging dataset, demonstrating, qualitatively and quantitatively, the advantages of our approach over previous work.
We experimentally show that our method produces reconstructions of higher accuracy and 3D stability, in particular in depth, than related work, also under difficult poses. 

%% file: sec/relatedwork.tex
\section{Related Work} 
\label{sec:relatedwork}
In the following, we focus on related work in the field of dense 3D human performance capture and do not review work on sparse 2D pose estimation.
%
%
\par \noindent\textbf{Capture using Parametric Models.}
Monocular human performance capture is an ill-posed problem due to its high dimensionality and ambiguity.
Low-dimensional parametric models can be employed as shape and deformation prior.
First, model-based approaches leverage a set of simple geometric primitives \cite{plankers01, sminchisescu03b, sigal04, metaxas93}.
Recent methods employ detailed statistical models learned from thousands of high-quality 3D scans \cite{anguelov05, hasler10, park08, pons15, loper15, kadlecek16, meekyoung17, weiss11, helten13, zhang14a, bogo15}.
Deep learning is widely used to obtain 2D and/or 3D joint detections or 3D vertex positions that can be used to inform model fitting \cite{yinghao17,lassner17, mehta17, bogo16, kolotouros19}.
An alternative is to regress model parameters directly \cite{kanazawa18, pavlakos18, kanazawa19}.
Beyond body shape and pose, recent models also include facial expressions and hand motion \cite{pavlakos19, xiang18, joo18, romero17} leading to very expressive reconstruction results.
Since parametric body models do not represent garments, variation in clothing cannot be reconstructed, and therefore many methods recover the naked body shape under clothing \cite{balan07a, bualan08, zhang17, yang16}.
The full geometry of the actor can be reconstructed by non-rigidly deforming the base parametric model to better fit the observations \cite{rhodin16b, alldieck18a, alldieck18b}.
But they can only model tight clothes such as T-shirts and pants, but not loose apparel which has a different topology than the body model, such as skirts.
To overcome this problem, ClothCap \cite{pons17} captures the body and clothing separately, but requires active multi-view setups.
Physics based simulations have recently been leveraged to constrain tracking (SimulCap~\cite{tao19}), or to learn a model of clothing on top of SMPL (TailorNet~\cite{patel20}).
%
Instead, our method is based on person-specific templates including clothes and employs deep learning to predict clothing deformation based on monocular video directly.

%
%
\par \noindent\textbf{Depth-based Template-free Capture.}
Most approaches based on parametric models ignore clothing.
The other side of the spectrum are prior-free approaches based on one or multiple depth sensors.
Capturing general non-rigidly deforming scenes \cite{slavcheva17,guo17}, even at real-time frame rates \cite{newcombe15, innmann2016, guo17}, is feasible, but only works reliably for small, controlled, and slow motions.
Higher robustness can be achieved by using higher frame rate sensors \cite{guo18, kowdle18} or multi-view setups \cite{ye12, dou16, orts16, dou17, zhang14b}.
Techniques that are specifically tailored to humans increase robustness \cite{yu17, yu18, ye14} by integrating a skeletal motion prior \cite{yu17} or a parametric model \cite{yu18, wei12}.
HybridFusion~\cite{zheng18} additionally incorporates a sparse set of inertial measurement units.
These fusion-style volumetric capture techniques \cite{huang16, allain15, leroy17, collet15, prada17} achieve impressive results, but do not establish a set of dense correspondences between all frames.
In addition, such depth-based methods do not directly generalize to our monocular setting, have a high power consumption, and typically do not work well under sunlight.
%
%
\par \noindent\textbf{Monocular Template-free Capture.}
Quite recently, fueled by the progress in deep learning, many template-free monocular reconstruction approaches have been proposed.
Due to their regular structure, many implicit reconstruction techniques \cite{varol18, zheng19} make use of uniform voxel grids.
DeepHuman~\cite{zheng19} combines a coarse scale volumetric reconstruction with a refinement network to add high-frequency details.
Multi-view CNNs can map 2D images to 3D volumetric fields enabling reconstruction of a clothed human body at arbitrary resolution \cite{zeng18}.
SiCloPe~\cite{natsume18} reconstructs a complete textured 3D model, including cloth, from a single image.
PIFu~\cite{saito19} regresses an implicit surface representation that locally aligns pixels with the global context of the corresponding 3D object.
Unlike voxel-based representations, this implicit per-pixel representation is more memory efficient.
These approaches have not been demonstrated to generalize well to strong articulation.
Furthermore, implicit approaches do not recover frame-to-frame correspondences which are of paramount importance for downstream applications, e.g., in augmented reality and video editing.
In contrast, our method is based on a mesh representation and can explicitly obtain the per-vertex correspondences over time while being slightly less general.
%
%
\par \noindent\textbf{Template-based Capture.}
An interesting trade-off between being template-free and relying on parametric models are approaches that only employ a template mesh as prior.
Historically, template-based human performance capture techniques exploit multi-view geometry to track the motion of a person \cite{starck07}.
Some systems also jointly reconstruct and obtain a foreground segmentation \cite{bray06, brox10, liu11, wu12}.
Given a sufficient number of multi-view images as input, some approaches \cite{carranza03, cagniart10, de08} align a personalized template model to the observations using non-rigid registration.
All the aforementioned methods require expensive multi-view setups and are not practical for consumer use.
Depth-based techniques enable template tracking from less cameras \cite{zollhoefer2014, ye12} and reduced motion models \cite{wu13, gall09, vlasic08, liu11} increase tracking robustness.
Recently, capturing 3D dense human body deformation just with a single RGB camera has been enabled \cite{xu18} and real-time performance has been achieved \cite{habermann19}.
However, their methods rely on expensive optimization leading either to very long per-frame computation times \cite{xu18} or the need for two graphics cards \cite{habermann19}.
Similar to them, our approach also employs a person-specific template mesh.
But differently, our method directly learns to predict the skeletal pose and the non-rigid surface deformations.
As shown by our experimental results, benefiting from our multi-view based self-supervision, our reconstruction accuracy significantly outperforms the existing methods.

%% file: sec/method.tex
\section{Method}
\label{sec:method}
%
%
\begin{figure*}[t]
	    \centering
	    \includegraphics[width=\textwidth]{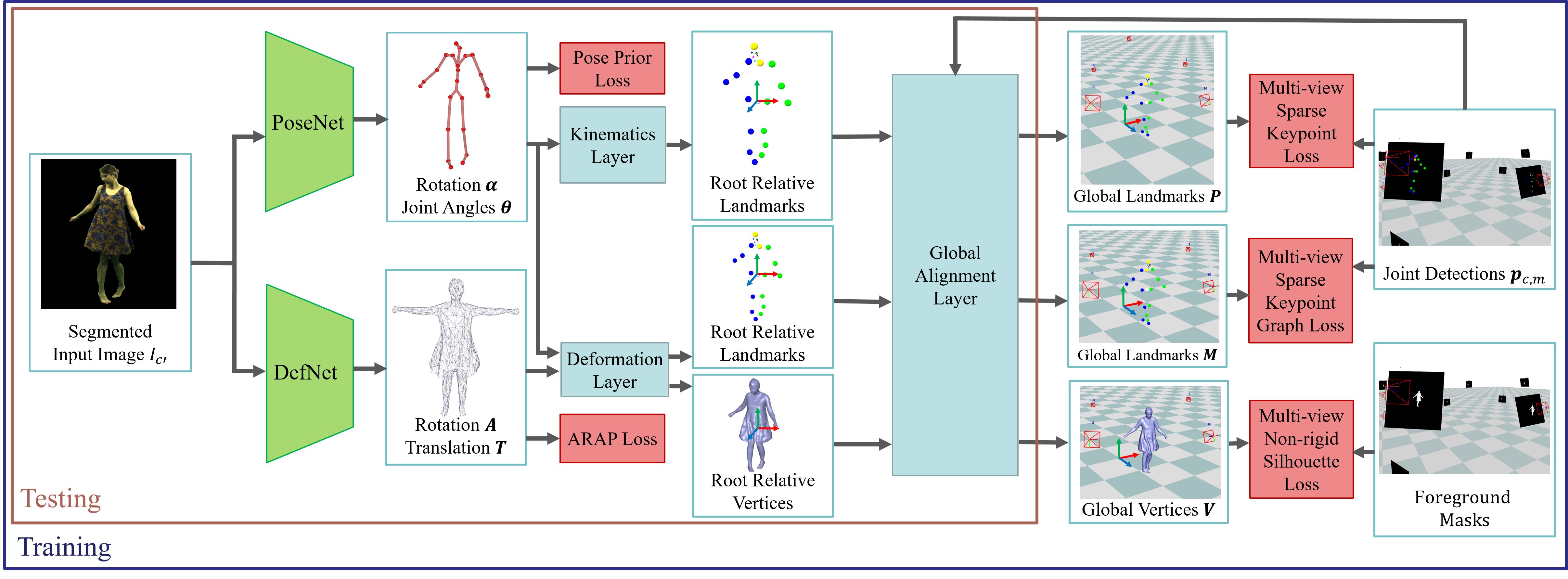} 
	    \caption
	    {
	    	Overview of our approach. 
	    	Our method takes a single segmented image as input. 
	    	First, our pose network, \textit{PoseNet}, is trained to predict the joint angles and the camera relative rotation using sparse multi-view 2D joint detections as weak supervision. 
	    	Second, the deformation network, \textit{DefNet}, is trained to regress embedded graph rotation and translation parameters to account for non-rigid deformations. 
	    	To train DefNet, multi-view 2D joint detections and silhouettes are used for supervision. 
    	}
	    \label{fig:overview}
    	\vspace{-4mm}
\end{figure*}
%
\input{sec/m1.tex}
%
\input{sec/m2.tex}
%
\input{sec/m3.tex}
\input{sec/m4.tex}

%% file: sec/m1.tex
%
%
\par 
Given a single RGB video of a moving human in general clothing, our goal is to capture the dense deforming surface of the full body.
This is achieved by training a neural network consisting of two components:
As illustrated in Fig.~\ref{fig:overview}, our pose network, \textit{PoseNet}, estimates the skeletal pose of the actor in the form of joint angles from a monocular image (Sec.~\ref{sec:poseNetwork}). 
Next, our deformation network, \textit{DefNet}, regresses the non-rigid deformation of the dense surface, which cannot be modeled by the skeletal motion, in the embedded deformation graph representation (Sec.~\ref{sec:deformationNetwork}).
To avoid generating dense 3D ground truth annotation, our network is trained in a weakly supervised manner.
To this end, we propose a fully differentiable human deformation and rendering model, which allows us to compare the rendering of the human body model to the 2D image evidence and back-propagate the losses.
For training, we first capture a video sequence in a calibrated multi-camera green screen studio (Sec.~\ref{sec:templateDataAndAcquisition}).
Note that our multi-view video is only used during training.
At test time we only require a single RGB video to perform dense non-rigid tracking.
%
%
\subsection{Template and Data Acquisition}
\label{sec:templateDataAndAcquisition}
%
%
\par \noindent\textbf{Character Model.}
Our method relies on a person-specific 3D template model.
We first scan the actor with a 3D scanner~\cite{treedys} to obtain the textured mesh.
Then, it is automatically rigged to a kinematic skeleton, which is parameterized with joint angles $\boldsymbol{\theta} \in \mathbb{R}^{27}$, the camera relative rotation $\boldsymbol{\alpha} \in \mathbb{R}^{3}$ and translation $\mathbf{t} \in \mathbb{R}^{3}$.
To model the non-rigid surface deformation, we automatically build an embedded deformation graph $\mathcal{G}$ with $K$ nodes following~\cite{sumner07}.
The nodes are parameterized with Euler angles $\mathbf{A} \in \mathbb{R}^{K \times 3}$ and translations $\mathbf{T} \in \mathbb{R}^{K \times 3}$.
Similar to \cite{habermann19}, we segment the mesh into different non-rigidity classes resulting in per-vertex rigidity weights $s_{i}$.
This allows us to model varying deformation behaviors of different surface materials, e.g. skin deforms less than clothing (see Eq.~\ref{eq:lossARAP}).
%
%
\par \noindent\textbf{Training Data.}
To acquire the training data, we record a multi-view video of the actor doing various actions in a calibrated multi-camera studio with green screen.
To provide weak supervision for the training, we first perform 2D pose detection on the sequences using OpenPose~\cite{cao17,cao18,simon17,wei16} and apply temporal filtering.
Then, we generate the foreground mask using color keying and compute the corresponding distance transform image $D_{f,c}$~\cite{borgefors86}, where $f \in [0,F]$ and $c \in [0,C]$ denote the frame index and camera index, respectively.
During training, we randomly sample one camera view $c'$ and frame $f'$ for which we crop the recorded image with a bounding box, based on the 2D joint detections.
The final training input image $I_{f',c'} \in \mathbb{R}^{256 \times 256 \times 3}$ is obtained by removing the background and augmenting the foreground with random brightness, hue, contrast and saturation changes.
For simplicity, we describe the operation on frame $f'$ and omit the subscript $f'$ in following equations.
%

%% file: sec/m2.tex
%
\subsection{Pose Network}
\label{sec:poseNetwork}
In our \textit{PoseNet}, we use ResNet50~\cite{he16} pretrained on ImageNet~\cite{deng09} as backbone and modify the last fully connected layer to output a vector containing the joint angles $\boldsymbol{\theta}$ and the camera relative root rotation $\boldsymbol{\alpha}$, given the input image $I_{c'}$. 
Since generating the ground truth for $\boldsymbol{\theta}$ and $\boldsymbol{\alpha}$ is a non-trivial task, we propose weakly supervised training based on fitting the skeleton to multi-view 2D joint detections.
%
%
\par \noindent\textbf{Kinematics Layer.}
To this end, we introduce a kinematics layer as the differentiable function that takes the joint angles $\boldsymbol{\theta}$ and the camera relative rotation $\boldsymbol{\alpha}$ and computes the positions $\mathbf{P}_{c'} \in \mathbb{R}^{M \times 3}$ of the $M$ 3D landmarks attached to the skeleton (17 body joints and 4 face landmarks). 
Note that $\mathbf{P}_{c'}$ lives in a camera-root-relative coordinate system.
In order to project the landmarks to other camera views, we need to transform $\mathbf{P}_{c'}$ to the world coordinate system:
\begin{equation}
\label{eq:global_transform}
    \mathbf{P}_m = \mathbf{R}_{c'}^{T} \mathbf{P}_{c',m} + \mathbf{t},
\end{equation}
where $\mathbf{R}_{c'}$ is the rotation matrix of the input camera $c'$ and $\mathbf{t}$ is the global translation of the skeleton.
%
%
\par \noindent\textbf{Global Alignment Layer.}
To obtain the global translation $\mathbf{t}$, we propose a global alignment layer that is attached to the kinematics layer.
It localizes our skeleton model in the world space, such that the globally rotated landmarks $\mathbf{R}_{c'}^{T} \mathbf{P}_{c',m}$ project onto the corresponding detections in all camera views.
This is done by minimizing the distance between the rotated landmarks $\mathbf{R}_{c'}^{T} \mathbf{P}_{c',m}$ and the corresponding rays cast from the camera origin $\mathbf{o}_c$ to the 2D joint detections:
\begin{equation} \label{eq:EGlobalAlign}
\sum_c \sum_m
\sigma_{c,m}
		\lVert
			(\mathbf{R}_{c'}^T \mathbf{P}_{c',m} + \mathbf{t} -\mathbf{o}_c) \times \mathbf{d}_{c,m}
		\rVert^2,
\end{equation}
where $\mathbf{d}_{c,m}$ is the direction of a ray from camera $c$ to the 2D joint detection $\mathbf{p}_{c,m}$ corresponding to landmark $m$:
\begin{equation}
\mathbf{d}_{c,m}= 
\frac
	{(\mathbf{E}_c^{-1} \tilde{\mathbf{p}}_{c,m})_{xyz} -\mathbf{o}_c }
	{\lVert(\mathbf{E}_c^{-1} \tilde{\mathbf{p}}_{c,m})_{xyz} -\mathbf{o}_c \rVert}
	\mathrm{.}
\end{equation}
Here, $\mathbf{E}_c \in \mathbb{R}^{4 \times 4}$ is the projection matrix of camera $c$ and $\tilde{\mathbf{p}}_{c,m} = (\mathbf{p}_{c,m},1,1)^T$.
Each point-to-line distance is weighted by the joint detection confidence $\sigma_{c,m}$, which is set to zero if below $0.4$. 
The minimization problem of Eq.~\ref{eq:EGlobalAlign} can be solved in closed form:
\begin{equation}\label{eq:globalalignop}
    \mathbf{t} = 
    \mathbf{W}^{-1}
	    \sum_{c,m}
	    	\mathbf{D}_{c,m}(\mathbf{R}_{c'}^T \mathbf{P}_{c',m} -\mathbf{o}_{c}) + \mathbf{o}_{c} - \mathbf{R}_{c'}^T \mathbf{P}_{c',m} 
	\mathrm{,}
\end{equation}
where 
\begin{equation}
\mathbf{W}
=
	\sum_c
	\sum_m 
	\mathbf{I}-\mathbf{D}_{c,m}
	\mathrm{.}
\end{equation}
Here, $\mathbf{I}$ is the $3\times3$ identity matrix and $\mathbf{D}_{c,m}=\mathbf{d}_{c,m}\mathbf{d}_{c,m}^T$.
Note that the operation in Eq.~\ref{eq:globalalignop} is differentiable with respect to the landmark position $\mathbf{P}_{c'}$.
%
%
\par \noindent\textbf{Sparse Keypoint Loss.}
Our 2D sparse keypoint loss for the \textit{PoseNet} can be expressed as
\begin{equation} \label{eq:lossKeypoint}
	\mathcal{L}_{\mathrm{kp}}(\mathbf{P}) = 
	\sum_c \sum_m
				\lambda_{m} \sigma_{c,m}
				\lVert 
					\pi_c\left(	
						\mathbf{P}_m
					\right)
					-
					\mathbf{p}_{c,m}
				\rVert^2 \mathrm{,}
\end{equation}
which ensures that each landmark projects onto the corresponding 2D joint detections $\mathbf{p}_{c,m}$ in all camera views.
Here, $\pi_c$ is the projection function of camera $c$ and $\sigma_{c,m}$ is the same as in Eq.~\ref{eq:EGlobalAlign}.
$\lambda_{m}$ is a kinematic chain-based hierarchical weight which varies during training for better convergence (see the supplementary material for details).
%
%
\par \noindent\textbf{Pose Prior Loss.}
To avoid unnatural poses, we impose a pose prior loss on the joint angles
\begin{equation} \label{eq:limit}
	\mathcal{L}_\mathrm{limit}(\boldsymbol{\theta}) = \sum_{i=1}^{27}{ \Psi( \boldsymbol{\theta}_i ) }
\end{equation}
\begin{equation} \label{eq:psi}
	\Psi(x)
	=
	\begin{cases}
		(x - \boldsymbol{\theta}_{\mathrm{max},i})^2,\text{ if } x > \boldsymbol{\theta}_{\mathrm{max},i}\\
		(\boldsymbol{\theta}_{\mathrm{min},i} - x)^2 \, ,\text{ if } x < \boldsymbol{\theta}_{\mathrm{min},i}\\
		0 \qquad \qquad \; \; \; , \text{ otherwise}
	\end{cases}\mathrm{,}
\end{equation}
that encourages that each joint angle $\boldsymbol{\theta}_i$ stays in a range $[\boldsymbol{\theta}_{\mathrm{min},i},\boldsymbol{\theta}_{\mathrm{max},i}]$ depending on the anatomic constraints.
%
%

%% file: sec/m3.tex
\subsection{Deformation Network}
\label{sec:deformationNetwork}
%
%
With the skeletal pose from \textit{PoseNet} alone, the non-rigid deformation of the skin and clothes cannot be fully explained.
Therefore, we disentangle the non-rigid deformation and the articulated skeletal motion.
\textit{DefNet} takes the input image $I_{c'}$ and regresses the non-rigid deformation parameterized with rotation angles $\mathbf{A}$ and translation vectors $\mathbf{T}$ of the nodes of the embedded deformation graph. 
\textit{DefNet} uses the same backbone architecture as \textit{PoseNet}, while the last fully connected layer outputs a $6K$-dimensional vector reshaped to match the dimensions of $\mathbf{A}$ and $\mathbf{T}$.
The weights of \textit{PoseNet} are fixed while training \textit{DefNet}.
Again, we do not use direct supervision on $\mathbf{A}$ and $\mathbf{T}$.
Instead, we propose a deformation layer with differentiable rendering and use multi-view silhouette-based weak supervision.
%
%
\par \noindent\textbf{Deformation Layer.}
The deformation layer takes $\mathbf{A}$ and $\mathbf{T}$ from \textit{DefNet} as input to non-rigidly deform the surface
\begin{equation} \label{eq:vnr}
	\mathbf{Y}_{i}= 
	\sum_{k \in \mathcal{N}_{\mathrm{vn}}(i)}
	w_{i,k}
	(
		R(\mathbf{A}_k)(\hat{\mathbf{V}}_i-\mathbf{G}_k) +\mathbf{G}_k + \mathbf{T}_k
	)
	\mathrm{.}
\end{equation}
Here, $\mathbf{Y}, \hat{\mathbf{V}} \in \mathbb{R}^{N \times 3}$ are the vertex positions of the deformed and undeformed template mesh, respectively.
$w_{i,k}$ are vertex-to-node weights, but in contrast to \cite{sumner07} we compute them based on geodesic distances. 
$\mathbf{G}\in \mathbb{R}^{K \times 3}$ are the node positions of the undeformed graph, $\mathcal{N}_{\mathrm{vn}}(i)$ is the set of nodes that influence vertex $i$, and $R(\cdot)$ is a function that converts the Euler angles to rotation matrices.
We further apply the skeletal pose on the deformed mesh vertices to obtain the vertex positions in the input camera space
\begin{equation} \label{eq:vi}
	\mathbf{V}_{\mathrm{c'},i} =
	\sum_{k \in \mathcal{N}_{\mathrm{vn}}(i)}
	w_{i,k}
	(
	R_{\mathrm{sk},k}(\boldsymbol{\theta}, \boldsymbol{\alpha}) \mathbf{Y}_{i} + t_{\mathrm{sk},k}(\boldsymbol{\theta}, \boldsymbol{\alpha})
	)
	\mathrm{,}
\end{equation}
where the node rotation $R_{\mathrm{sk},k}$ and translation $t_{\mathrm{sk},k}$ are derived from the pose parameters using dual quaternion skinning~\cite{kavan07}.
Eq.~\ref{eq:vnr} and Eq.~\ref{eq:vi} are differentiable with respect to pose and graph parameters.
Thus, our layer can be integrated in the learning framework and gradients can be propagated to \textit{DefNet}.
So far, $\mathbf{V}_{\mathrm{c'},i}$ is still rotated relative to the camera $c'$ and located around the origin.
To bring them to global space, we apply the inverse camera rotation and the global translation, defined in Eq.~\ref{eq:globalalignop}, $\mathbf{V}_i = \mathbf{R}_{c'}^{T} \mathbf{V}_{\mathrm{c'},i} + \mathbf{t}$.
%
%
\par \noindent\textbf{Non-rigid Silhouette Loss.}
This loss encourages that the non-rigidly deformed mesh matches the multi-view silhouettes in all camera views.
It can be formulated using the distance transform representation~\cite{borgefors86}
\begin{equation} \label{eq:lossSil}
	\mathcal{L}_{\mathrm{sil}}(	\mathbf{V}) = 
		\sum_{c} \sum_{i \in \mathcal{B}_c} 
				\rho_{c,i}
				\|D_{c}\left(
						\pi_c\left(	
							\mathbf{V}_i
						\right)
					\right)
			 	\|^2\mathrm{.}
\end{equation}
Here, $\mathcal{B}_c $ is the set of vertices that lie on the boundary when the deformed 3D mesh is projected onto the distance transform image $D_{c}$ of camera $c$, and $\rho_{c,i}$ is a directional weighting~\cite{habermann19} that guides the gradient in $D_{c}$.
The silhouette loss ensures that the boundary vertices project onto the zero-set of the distance transform, \textit{i.e.}, the foreground silhouette.
%
%
\par \noindent\textbf{Sparse Keypoint Graph Loss.}
Only using the silhouette loss can lead to wrong mesh-to-image assignments, especially for highly articulated motions.
To this end, we use a sparse keypoint loss to constrain the mesh deformation, which is similar to the keypoint loss for \textit{PoseNet} in Eq.~\ref{eq:lossKeypoint}
\begin{multline} \label{eq:keypointGraphLoss}
\mathcal{L}_{\mathrm{kpg}}(\mathbf{M}) = 
\sum_c \sum_m
\sigma_{c,m}
\lVert 
\pi_c\left(	
\mathbf{M}_m
\right)
-
\mathbf{p}_{c,m}
\rVert^2
\mathrm{.}
\end{multline}
Differently from Eq.~\ref{eq:lossKeypoint}, the deformed and posed landmarks $\mathbf{M}$ are derived from the embedded deformation graph.
To this end, we can deform and pose the canonical landmark positions by attaching them to its closest graph node $g$ in canonical pose with weight $w_{m,g}=1.0$. 
Landmarks can then be deformed according to Eq.~\ref{eq:vnr}, ~\ref{eq:vi}, resulting in $\mathbf{M}_{c'}$ which is brought to global space via $\mathbf{M}_m = \mathbf{R}_{c'}^{T} \mathbf{M}_{\mathrm{c'},m} + \mathbf{t}$.
%
%
\par \noindent\textbf{As-rigid-as-possible Loss.}
To enforce local smoothness of the surface, we impose an as-rigid-as-possible loss~\cite{sorkine07}
\begin{equation} \label{eq:lossARAP}
\mathcal{L}_\mathrm{arap}(\mathbf{A},\mathbf{T}) = 
\sum_k \sum_{l \in \mathcal{N}_\mathrm{n}(k)}
u_{k,l}
\lVert 
d_{k,l}(\mathbf{A},\mathbf{T}) 
\rVert_1,
\end{equation}
where
$$
d_{k,l}(\mathbf{A},\mathbf{T})\! \!= \!\!
R(\mathbf{A}_k) (\mathbf{G}_l - \mathbf{G}_k) + \mathbf{T}_k + \mathbf{G}_k
- (\mathbf{G}_l + \mathbf{T}_l).
$$
$\mathcal{N}_\mathrm{n}(k)$ is the set of indices of the neighbors of node $k$.
In contrast to \cite{sorkine07}, we propose weighting factors $u_{k,l}$ that influence the rigidity of respective parts of the graph.
We derive $u_{k,l}$ by averaging all per-vertex rigidity weights $s_{i}$ \cite{habermann19} for all vertices (see Sec.~\ref{sec:templateDataAndAcquisition}), which are connected to node $k$ or $l$.
Thus, the mesh can deform either less or more depending on the surface material. 
For example, graph nodes that are mostly connected to vertices on a skirt can deform more freely than nodes that are mainly connected to vertices on the skin.

%% file: sec/m4.tex
\subsection{In-the-wild Domain Adaptation}
\label{sec:domainAdaptation}
Since our training set is captured in a green screen studio and our test set is captured in the wild, there is a significant domain gap between them, due to different lighting conditions and camera response functions.
To improve the performance of our method on in-the-wild images, we fine-tune our networks on the monocular test images for a small number of iterations using the same 2D keypoint and silhouette losses as before, \emph{but only on a single view}.
This drastically improves the performance at test time as shown in the supplemental material.

%% file: sec/evaluation.tex
\section{Results}
\label{sec:evaluation}
\begin{figure*}[t]
	\begin{center}
		\includegraphics[width=\linewidth]{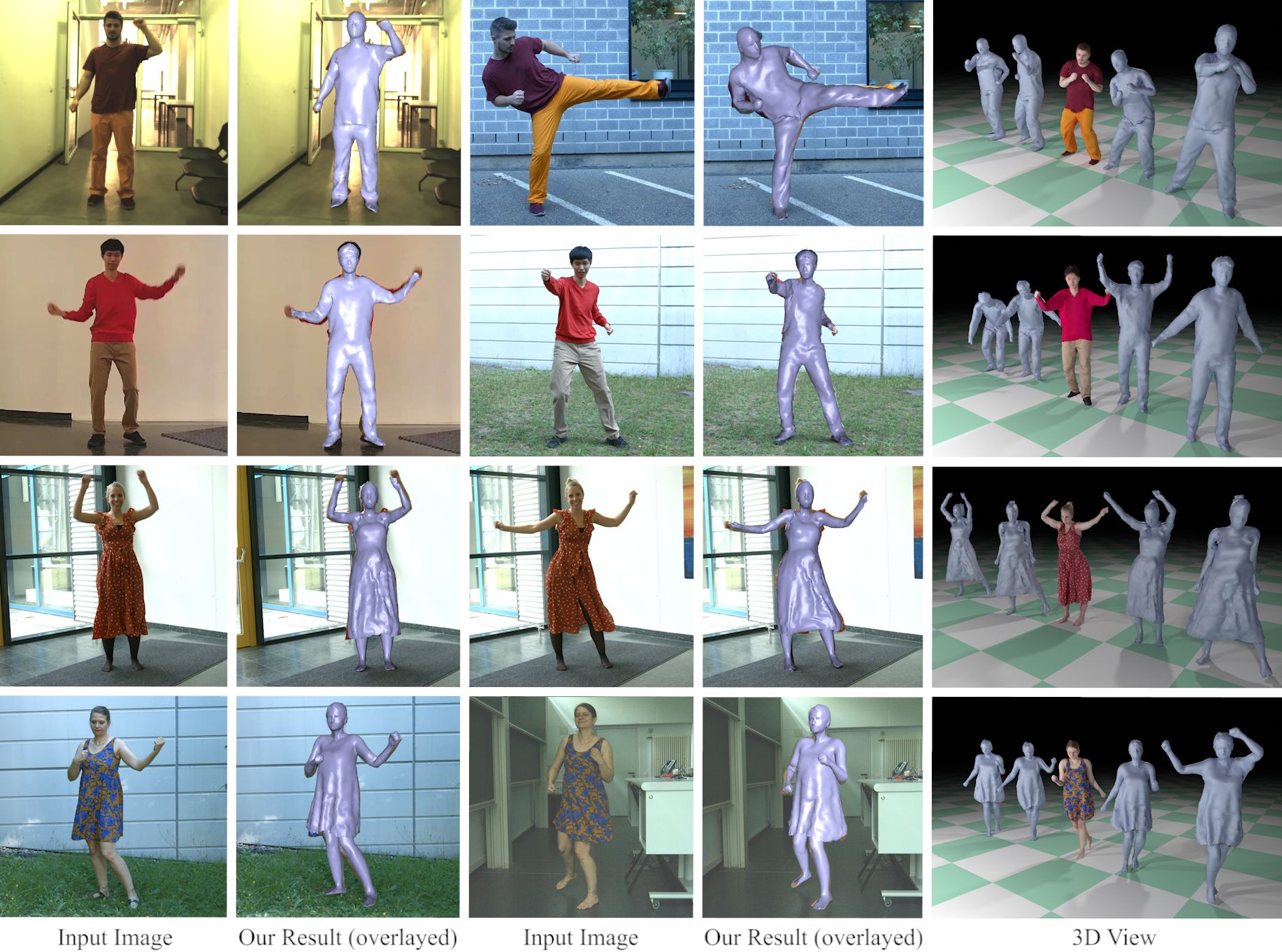}
	\end{center}
	\vspace{-2mm}
	\caption
	{
		Qualitative results. 
		Each row shows results for a different person with varying types of apparel. 
		We visualize input frames and our reconstruction overlayed to the corresponding frame.
		Note that our results precisely overlay to the input.
		Further, we show our reconstructions from a virtual 3D viewpoint.
		Note that they also look plausible in 3D.
	}
	\label{fig:qualitative}
	\vspace{-4mm}
\end{figure*}
All our experiments were performed on a machine with an NVIDIA Tesla V100 GPU.
A forward pass of our method takes around 50ms, which breaks down to 25ms for \textit{PoseNet} and 25ms for \textit{DefNet}.
During testing, we use the off-the-shelf video segmentation method of~\cite{caelles17} to remove the background in the input image.
Our method requires OpenPose's 2D joint detections~\cite{cao17,cao18,simon17,wei16} as input during testing to crop the frames and to obtain the 3D global translation with our global alignment layer.
Finally, we temporally smooth the output mesh vertices with a Gaussian kernel of size 5 frames.
%
%
\par
\begin{figure}[t]
	\begin{center}
		\includegraphics[width=\linewidth]{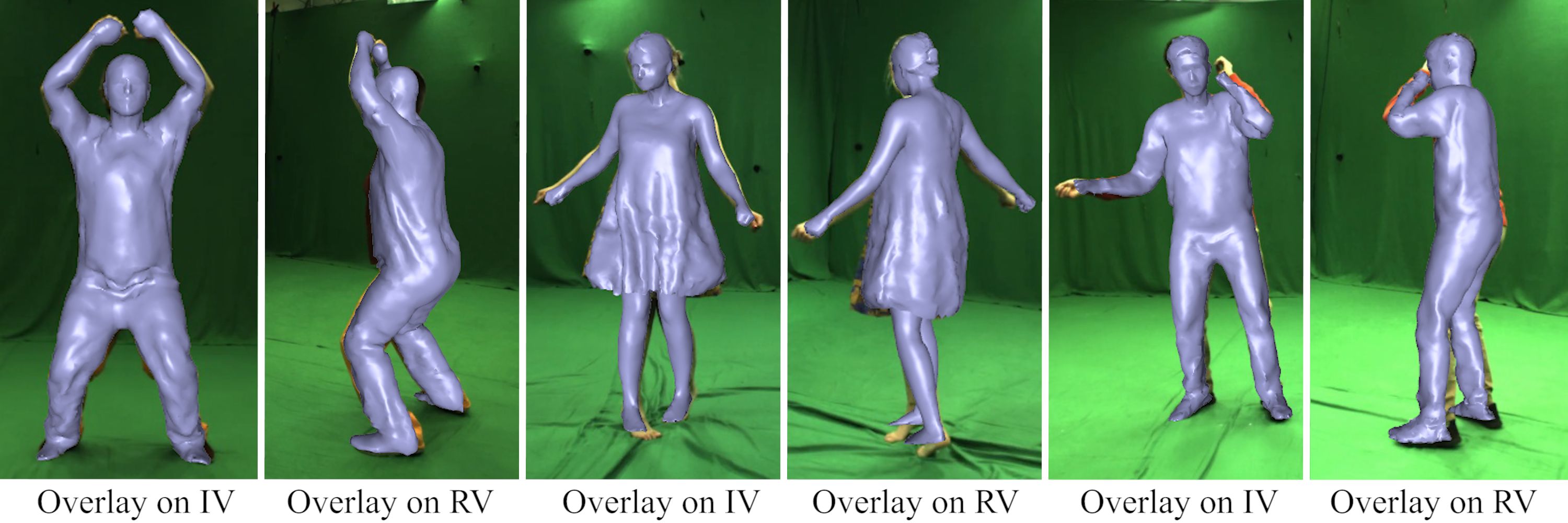}
	\end{center}
	\vspace{-2mm}
	\caption
	{
		Results on our evaluation sequences where input views (IV) and reference views (RV) are available.
		Note that our reconstruction also precisely overlays on RV even though they are not used for tracking.
	}
	\label{fig:referenceview}
	\vspace{-4mm}
\end{figure}

 \noindent \textbf{Dataset.}
We evaluate our approach on 4 subjects (\textit{S1} to \textit{S4}) with varying types of apparel.
For qualitative evaluation, we recorded 13 in-the-wild sequences in different indoor and outdoor environments shown in Fig.~\ref{fig:qualitative}.
For quantitative evaluation, we captured 4 sequences in a calibrated multi-camera green screen studio (see Fig.~\ref{fig:referenceview}), for which we computed the ground truth 3D joint locations using the multi-view motion capture software, The Captury~\cite{captury}, and we use a color keying algorithm for ground truth foreground segmentation.
All sequences contain a large variety of motions, ranging from simple ones like walking up to more difficult ones like fast dancing or baseball pitching. 
We will release the dataset for future research.
%
%
%
%
\par \noindent \textbf{Qualitative Comparisons.}
Fig.~\ref{fig:qualitative} shows our qualitative results on in-the-wild test sequences with various clothing styles, poses and environments.
Our reconstructions not only precisely overlay with the input images, but also look plausible from arbitrary 3D view points.
In Fig.~\ref{fig:qualcomp}, we qualitatively compare our approach to the related human capture and reconstruction methods~\cite{kanazawa18, habermann19, saito19, zheng19}.
In terms of the shape representation, our method is most closely related to LiveCap~\cite{habermann19} that also uses a person-specific template.
Since they non-rigidly fit the template only to the monocular input view, their results do not faithfully depict the deformation in other view points.
Further, their pose estimation severely suffers from the monocular ambiguities, whereas our pose results are more robust and accurate (see supplemental video).
Comparing to the other three methods~\cite{kanazawa18, saito19, zheng19} that are trained for general subjects, our approach has the following advantages:
First, our method recovers the non-rigid deformations of humans in general clothes whereas the parametric model-based approaches \cite{kanazawa18,kanazawa19} only recover naked body shape.
Second, our method directly provides surface correspondences over time which is important for AR/VR applications (see supplemental video).
In contrast, the results of implicit representation-based methods, PIFu~\cite{saito19} and DeepHuman~\cite{zheng19}, lack temporal surface correspondences and do not preserve the skeletal structure of the human body, \textit{i.e.}, they often exhibit missing arms and disconnected geometry.
Furthermore, DeepHuman~\cite{zheng19} only recovers a coarse shape in combination with a normal image of the input view, while our method can recover medium-level detailed geometry that looks plausible from all views.
Last but not least, all these existing methods have problems when overlaying their reconstructions on the reference view, even though some of the methods show a very good overlay on the input view.
In contrast, our approach reconstructs accurate 3D geometry, and therefore, our results can precisely overlay on the reference views (also see Fig.~\ref{fig:referenceview}).
\newcommand{\bmethodlabel}[2]{\scriptsize {\hspace{#1} {#2}}}	
\begin{figure}[t]
	\scriptsize
	\begin{center}
		\includegraphics[width=\linewidth]{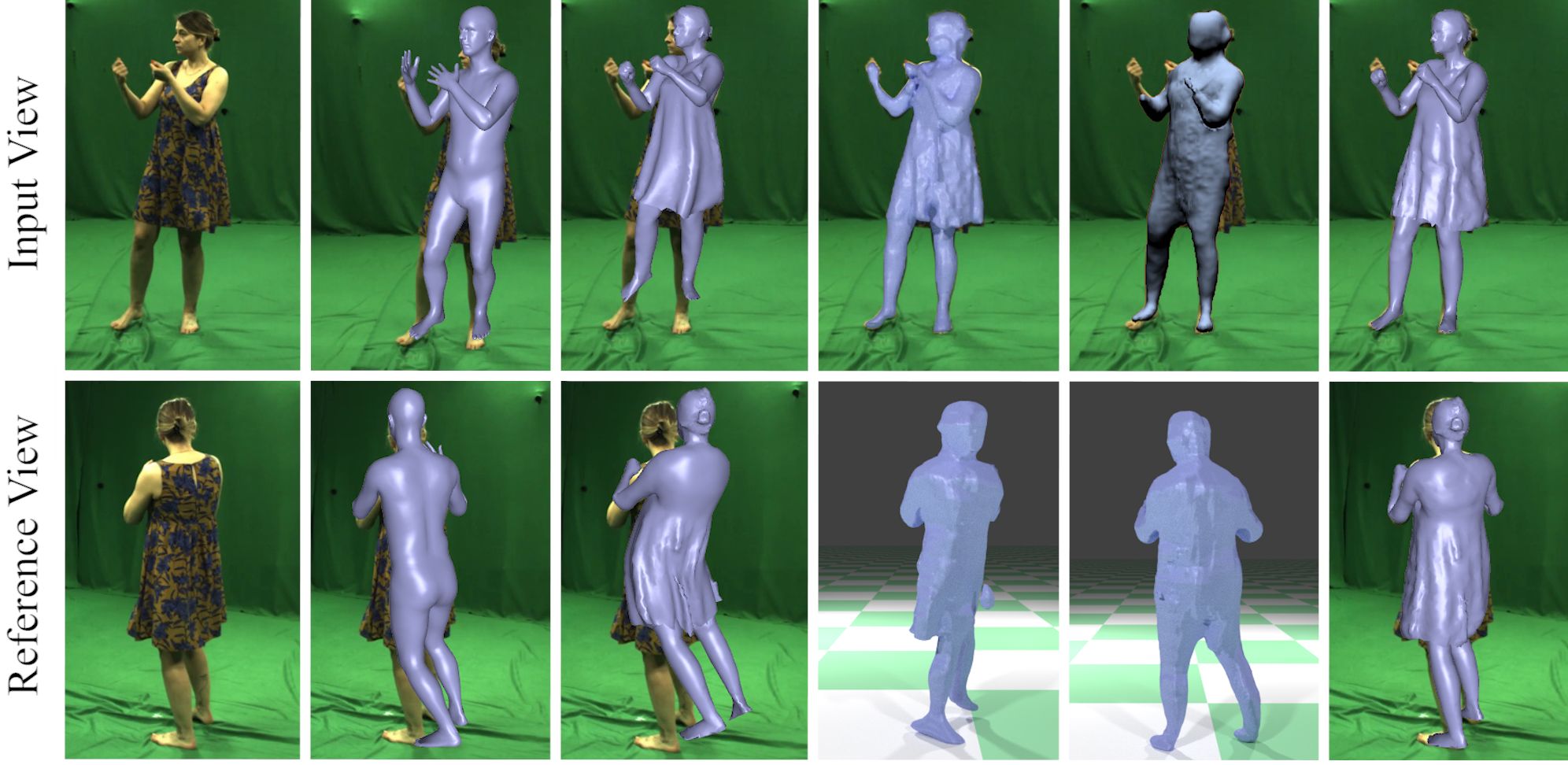}
	\end{center}
	\vspace{-0.4cm}
	\bmethodlabel{0.6cm}{Input}
	\bmethodlabel{0.4cm}{HMR~\cite{kanazawa18}}
	\bmethodlabel{0.1cm}{LiveCap~\cite{habermann19}}
	\bmethodlabel{0.1cm}{PIFu~\cite{saito19}}
	\bmethodlabel{0.01cm}{DeepHuman~\cite{zheng19}}
	\bmethodlabel{0.1cm}{\textbf{Ours}}
	\caption
	{
		Qualitative comparison to other methods \cite{kanazawa18,habermann19,saito19,zheng19}. 
		Note that our results overlay more accurately to the input view and also look more plausible from a reference view that was not used for tracking.
		Ground truth global translation is used to match the reference view for the results of~\cite{kanazawa18,habermann19}.
		Since PIFu~\cite{saito19} and DeepHuman~\cite{zheng19} output meshes with varying topology in a canonical volume without an attached root, it is not possible to apply the ground truth translation and therefore we show the reference view without overlay.
	}
	\label{fig:qualcomp}
	\vspace{-5mm}
\end{figure}
%
%
\par \noindent \textbf{Skeletal Pose Accuracy.}
We quantitatively compare our pose results (output of \textit{PoseNet}) to existing pose estimation methods on \textit{S1} and \textit{S4}.
To account for different types of apparel, we choose \textit{S1} wearing trousers and a T-shirt and \textit{S4} wearing a short dress.
We rescale the bone length for all methods to the ground truth and evaluate the following metrics on the 14 commonly used joints~\cite{mehta17} for every 10th frame:
1) We evaluate the root joint position error or global localization error (\textit{GLE}) to measure how good the skeleton is placed in global 3D space.
Note that \textit{GLE} can only be evaluated for LiveCap~\cite{habermann19} and ours, since other methods only produce up-to-scale depth.
2) To evaluate the accuracy of the pose estimation, we report the 3D percentage of correct keypoints (3DPCK) with a threshold of $150mm$ of the root aligned poses and the area under the 3DPCK curve (AUC).
3) To factor out the errors in the global rotation, we also report the mean per joint position error (MPJPE) after Procrustes alignment.
We compare our approach against the state-of-the-art pose estimation approaches including VNect~\cite{mehta17}, HMR~\cite{kanazawa18}, HMMR~\cite{kanazawa19}, and LiveCap~\cite{habermann19}.
We also compare to a multi-view baseline approach (\textit{MVBL}), where we use our differentiable skeleton model in an optimization framework to solve for the pose per frame using the proposed multi-view losses.
We can see from Tab.~\ref{tab:ablationJoint} that our approach outperforms the related monocular methods in all metrics by a large margin and is even close to \textit{MVBL} although our method only takes a single image as input.
We further compare to VNect~\cite{mehta17} fine-tuned on our training images for \textit{S1}.
To this end, we compute the 3D joint position using The Captury~\cite{captury} to provide ground truth supervision for VNect.
On the evaluation sequence for \textit{S1}, the fine-tuned VNect achieved 95.66\% 3DPCK, 52.13\% AUC and 47.16$mm$ MPJPE.
This shows our weakly supervised approach yields comparable or better results than supervised methods in the person-specific setting.
However, our approach does not require 3D ground truth annotation that is difficult to obtain, even for only sparse keypoints, let alone the dense surfaces.
\begin{table}
	\scriptsize
	\begin{center}
		\begin{tabular}{|c|c|c|c|c|}
			\hline
			\multicolumn{5}{|c|}{\textit{MPJPE/GLE (in mm) and 3DPCK/AUC (in \%) on S1}} \\
			\hline
			\textbf{Method}                            & \textbf{GLE}$\downarrow$		& \textbf{3DPCK}$\uparrow$	& \textbf{AUC}$\uparrow$	 & \textbf{MPJPE}$\downarrow$		\\
			\hline
			VNect~\cite{mehta17} 			           & -   		  		& 66.06  	  		& 28.02			& 77.19		   		\\
			HMR~\cite{kanazawa18}		   	           & -      	  		& 82.39	  			& 43.61		    & 72.61				\\
			HMMR~\cite{kanazawa19}		   	           & -  	  	        & 87.48	  			& 45.33		    & 72.40				\\
			LiveCap~\cite{habermann19} 	               & 317.01		  		& 71.13	  			& 37.90	        & 92.84			 	\\
			Ours 									   & \textbf{91.08}		  		& \textbf{98.43}     		& \textbf{58.71}		    & \textbf{49.11}		      	\\
			\hline
			MVBL                                       & 76.03	      		& 99.17	  			& 57.79		    & 45.44				\\
			\hline
		\end{tabular}
	\end{center}
	\vspace{-0.6cm}
	\begin{center}
		\begin{tabular}{|c|c|c|c|c|}
			\hline
			\multicolumn{5}{|c|}{\textit{MPJPE/GLE (in mm) and 3DPCK/AUC (in \%) on S4}} \\
			\hline
			\textbf{Method}                            & \textbf{GLE}$\downarrow$		& \textbf{3DPCK}$\uparrow$	& \textbf{AUC}$\uparrow$	 & \textbf{MPJPE}$\downarrow$		\\
			\hline
			VNect~\cite{mehta17} 			               				& -     		  		& 82.06  	  			& 42.73			& 72.62		   		\\
			HMR~\cite{kanazawa18}		   	             				& -     		  		& 86.88  	  			& 43.91			& 73.63		   		\\
			HMMR~\cite{kanazawa19}		   	             				& -		  		        & 82.80  	  			& 41.18			& 77.41		   		\\
			LiveCap~\cite{habermann19} 	               					& 248.67		  		& 75.11  	  			& 37.35			& 83.48		   	    \\
			Ours 									               	    & \textbf{96.56}		  			& \textbf{96.74}  	  			& \textbf{59.25}			& \textbf{45.40}		   		\\
			\hline
			MVBL                                                    	& 75.82		  			& 96.20  	  			& 57.27			& 45.12		   		\\
			\hline
		\end{tabular}
	\end{center}
	\caption
	{
		Skeletal pose accuracy.
		Note that we are consistently better than other monocular approaches.
		Moreover, we are even close to the multi-view baseline.
	}
	\label{tab:joint}
	\vspace{-2mm}
\end{table}
%
%
\par \noindent \textbf{Surface Reconstruction Accuracy.}
To evaluate the accuracy of the regressed non-rigid deformations, we compute the intersection over union (IoU) between the ground truth foreground masks and the 2D projection of the estimated shape on \textit{S1} and \textit{S4} for every 100th frame.
We evaluate the IoU on \textit{all views}, on \textit{all views expect the input view}, and on the \textit{input view} which we refer to as \textit{AMVIoU}, \textit{RVIoU} and \textit{SVIoU}, respectively.
To factor out the errors in global localization, we apply the ground truth translation to the reconstructed geometries.
For DeepHuman~\cite{zheng19} and PIFu~\cite{saito19}, we cannot report the \textit{AMVIoU} and \textit{RVIoU}, since we cannot overlay their results on reference views as discussed before.
Further, PIFu~\cite{saito19} by design achieves perfect overlay on the input view, since they regress the depth for each foreground pixel.
However, their reconstruction does not reflect the true 3D geometry (see Fig.~\ref{fig:qualcomp}).
Therefore, it is meaningless to report their \textit{SVIoU}.
Similarly, DeepHuman~\cite{zheng19} achieves high \textit{SVIoU}, due to their volumetric representation.
But their results are often wrong, when looking from side views.
In contrast, our method consistently outperforms all other approaches in terms of \textit{AMVIoU} and \textit{RVIoU}, which shows the high accuracy of our method in recovering the 3D geometry.
Further, we are again close to the multi-view baseline.
\begin{table}
	\scriptsize
	\begin{center}
		\begin{tabular}{|c|c|c|c|}
			\hline
			\multicolumn{4}{|c|}{\textit{AMVIoU, RVIoU, and SVIoU (in \%) on S1 sequence}} \\
			\hline
			\textbf{Method}                                       	& \textbf{AMVIoU}$\uparrow$           	& \textbf{RVIoU}$\uparrow$     			& \textbf{SVIoU}$\uparrow$     		\\
			\hline
			HMR~\cite{kanazawa18}		                   			& 62.25		                          	& 61.7                       			& 68.85    							\\
			HMMR~\cite{kanazawa19}		                   			& 65.98 	                          	& 65.58                       			& 70.77   							\\
			LiveCap~\cite{habermann19} 	                 			& 56.02	                          		& 54.21                       			& 77.75	       	  					\\
			DeepHuman~\cite{zheng19} 		            			& -		                                & -                         			& \textbf{91.57}				      			\\
			Ours 									                & \textbf{87.2}	                      	 		& \textbf{87.03}                         		& 89.26      						\\
			\hline
			MVBL                                                    & 91.74		                       		& 91.72                         		& 92.02								\\
			\hline
		\end{tabular}
	\end{center}
	\vspace{-0.6cm}
	\begin{center}
		\begin{tabular}{|c|c|c|c|}
			\hline
			\multicolumn{4}{|c|}{\textit{AMVIoU, RVIoU, and SVIoU (in \%) on S4 sequence}} \\
			\hline
			\textbf{Method}                                 		& \textbf{AMVIoU}$\uparrow$            & \textbf{RVIoU}$\uparrow$     			& \textbf{SVIoU}$\uparrow$     	   	\\
			\hline
			HMR~\cite{kanazawa18}		              				& 65.1		                           & 64.66                         			& 70.84		                		\\
			HMMR~\cite{kanazawa19}		              				& 63.79		                           & 63.29                         			& 70.23		                		\\
			LiveCap~\cite{habermann19} 	            				& 59.96	                               & 59.02                         			& 72.16		                   		\\
			DeepHuman~\cite{zheng19} 		        				& -		                               & -                                 		& 84.15				                \\
			Ours 									               	& \textbf{82.53}		                           & \textbf{82.22}                          		& \textbf{86.66}			               		\\
			\hline
			MVBL                                                  	& 88.14		                           & 88.03                          		& 89.66				 	            \\
			\hline
		\end{tabular}
	\end{center}
	\caption
	{
		Surface deformation accuracy.
		We outperform all other monocular methods and are even close to the multi-view baseline.
	}
	\label{tab:surface}
	\vspace{-4mm}
\end{table}
%
%
\par
\begin{figure}[t]
	\begin{center}
		\includegraphics[width=\linewidth]{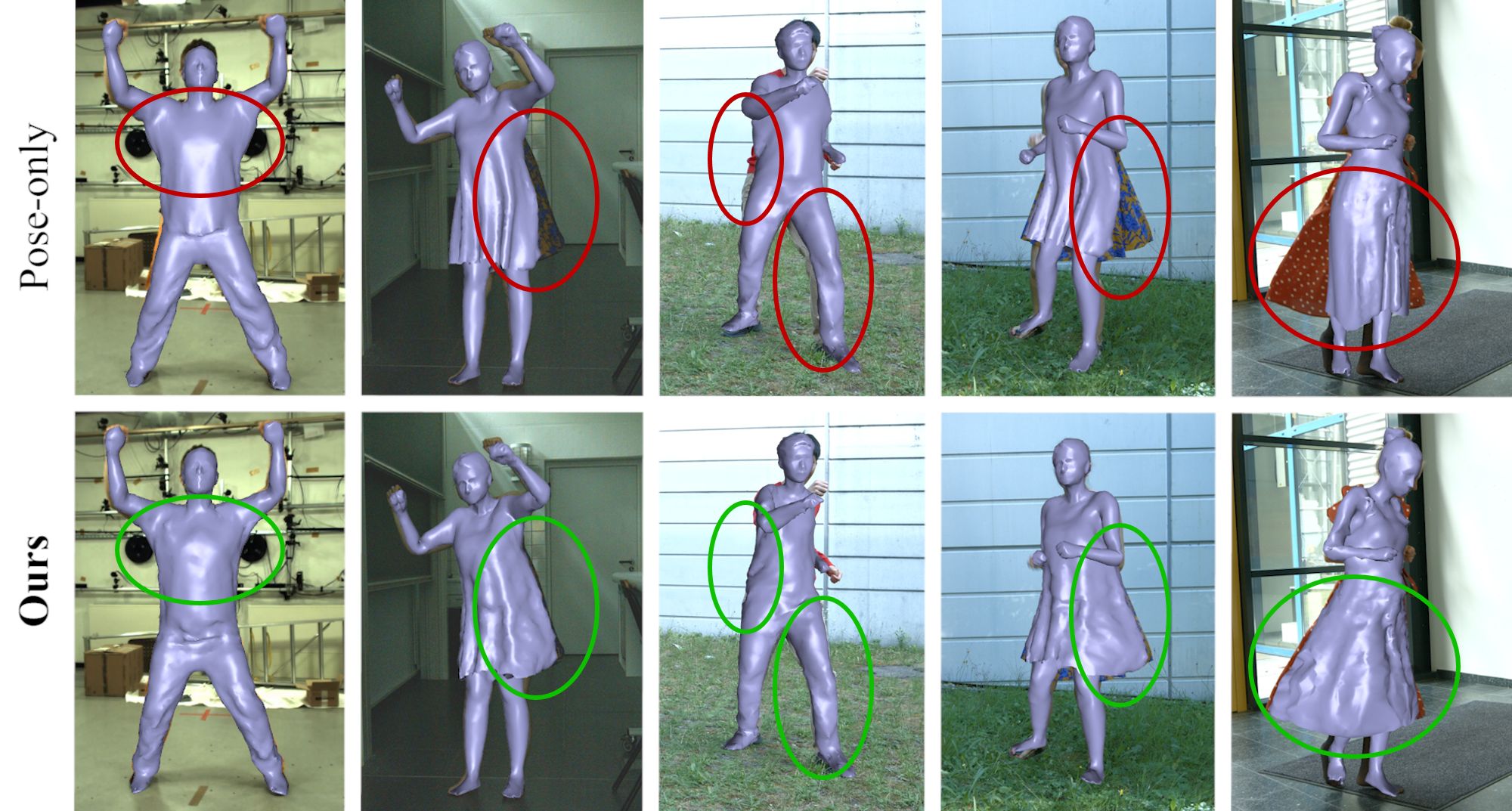}
	\end{center}
	\vspace{-4mm}
	\caption
	{
		\textit{PoseNet} + \textit{DefNet} vs. \textit{PoseNet-only}.
		\textit{DefNet} can deform the template to accurately match the input, especially for loose clothing.
		In addition, \textit{DefNet} also corrects slight errors in the pose and typical skinning artifacts.
	}
	\label{fig:nrvspose}
	\vspace{-2mm}
\end{figure}

 \noindent \textbf{Ablation Study.}
To evaluate the importance of the number of cameras, the number of training images, and our \textit{DefNet}, we performed an ablation study on \textit{S4} in Tab.~\ref{tab:ablationJoint}. 
1) In the first group of Tab.~\ref{tab:ablationJoint}, we train our networks with supervision using 1 to 7 views.
We can see that adding more views consistently improves the quality of the estimated poses and deformations.
The most significant improvement is from one to two cameras.
This is not surprising, since the single camera settings is inherently ambiguous.
2) In the second group of Tab.~\ref{tab:ablationJoint}, we reduce the training data to 1/2 and 1/4.
We can see that the more frames with different poses and deformations are seen during training, the better the reconstruction quality is.
This is expected since a larger number of frames may better sample the possible space of poses and deformations.
3) In the third group of Tab.~\ref{tab:ablationJoint}, we evaluate the \textit{AMVIoU} on the template mesh animated with the results of \textit{PoseNet}, which we refer to as \textit{PoseNet-only}.
One can see that on average, the \textit{AMVIoU} is improved by around 4\%.
Since most non-rigid deformations rather happen locally, the difference is visually even more significant as shown in Fig.~\ref{fig:nrvspose}.
Especially, the skirt is correctly deformed according to the input image whereas the \textit{PoseNet-only} result cannot fit the input due to the limitation of skinning.
\begin{table}
	\scriptsize
	\begin{center}
		\begin{tabular}{|c|c|c|}
			\hline
			\multicolumn{3}{|c|}{\textit{3DPCK and AMVIoU (in \%) on S4 sequence}} \\
			\hline
			\textbf{Method}                                     & \textbf{3DPCK}$\uparrow$   	& \textbf{AMVIoU}$\uparrow$    	\\
			\hline
			1 camera view 									        & 62.11		       			& 65.11		    				\\
			2 camera views 									        & 93.52	           			& 78.44	   						\\
			3 camera views 									        & 94.70		           		& 79.75		   					\\
			7 camera views 										    & 95.95		           		& 81.73							\\
			\hline
			6500 frames 									   	& 85.19		        	 	& 73.41		  					\\
			13000 frames 									   	& 92.25		        		& 78.97		     				\\
			\hline
			PoseNet-only                                        & 96.74	            		& 78.51		   					\\
			Ours(14 views, 26000 frames) 				  			& \textbf{96.74}		   	& \textbf{82.53}				\\
			\hline
		\end{tabular}
	\end{center}
	\caption
	{
		Ablation study.
		We evaluate the number of cameras and the number of frames used during training in terms of the \textit{3DPCK} and \textit{AMVIoU} metrics.	
		Adding more cameras and frames consistently improves the quality of reconstruction.
		Further, \textit{DefNet} improves the \textit{AMVIoU} compared to pure pose estimation.
	}
	\label{tab:ablationJoint}
	\vspace{-2mm}
\end{table}

%% file: sec/conclusion.tex
\section{Conclusion}
\label{sec:conclusion}
We have presented a learning-based approach for monocular dense human performance capture using only weak multi-view supervision. 
In contrast to existing methods, our approach directly regresses poses and surface deformations from neural networks, produces temporal surface correspondences, preserves the skeletal structure of the human body, and can handle loose clothes.
Our qualitative and quantitative results in different scenarios show that our method produces more accurate 3D reconstruction of pose and non-rigid deformation than existing methods.
In the future, we plan to incorporate hands and the face to our mesh representation to enable joint tracking of body, facial expressions and hand gestures.
We are also interested in physically more correct multi-layered representations to model the garments even more realistically.